\documentclass{article} %

\usepackage[final]{neurips_2024}

\usepackage{amsmath,amsfonts,bm}

\def\eqref#1{equation~\ref{#1}}

\def\1{\bm{1}}

\DeclareMathAlphabet{\mathsfit}{\encodingdefault}{\sfdefault}{m}{sl}
\SetMathAlphabet{\mathsfit}{bold}{\encodingdefault}{\sfdefault}{bx}{n}

\usepackage[utf8]{inputenc} %
\usepackage[T1]{fontenc}    %
\usepackage{hyperref}       %
\usepackage{url}            %
\usepackage{booktabs}       %
\usepackage{amsfonts}       %
\usepackage{nicefrac}       %
\usepackage{microtype}      %
\usepackage{xcolor}         %

\usepackage{tcolorbox}
\usepackage{placeins}
\usepackage{booktabs}
\usepackage{lscape}
\usepackage{wrapfig}

\title{Transfer Learning for Finetuning \\ Large Language Models 
}

\author{%
Tobias Strangmann\textsuperscript{1}, Lennart Purucker\textsuperscript{1}, Jörg K.H. Franke\textsuperscript{1}, Ivo Rapant\textsuperscript{1}, \\
\textbf{Fabio Ferreira\textsuperscript{1},\ Frank Hutter\textsuperscript{1,2}}
\vspace*{2mm}\\
\textsuperscript{1}University of Freiburg, \textsuperscript{2}ELLIS Institute Tübingen\\
}

\begin{document}

\maketitle

\begin{abstract}
As the landscape of large language models expands, efficiently finetuning for specific tasks becomes increasingly crucial.
At the same time, the landscape of parameter-efficient finetuning methods rapidly expands. 
Consequently, practitioners face a multitude of complex choices when searching for an optimal finetuning pipeline for large language models.  
To reduce the complexity for practitioners, we investigate transfer learning for finetuning large language models and aim to transfer knowledge about configurations from related finetuning tasks to a new task.
In this work, we transfer learn finetuning by meta-learning performance and cost surrogate models for grey-box meta-optimization from a new meta-dataset.
Counter-intuitively, we propose to rely only on transfer learning for new datasets. 
Thus, we do not use task-specific Bayesian optimization but prioritize knowledge transferred from related tasks over task-specific feedback.
We evaluate our method on eight synthetic question-answer datasets and a meta-dataset consisting of 1,800 runs of finetuning Microsoft's Phi-3.
Our transfer learning is superior to zero-shot, default finetuning, and meta-optimization baselines.
Our results demonstrate the transferability of finetuning to adapt large language models more effectively. 
\end{abstract}

\section{Introduction}
The landscape of large language models (LLMs) rapidly expands to a zoo of models \citep{gemma_2024,abdin2024phi3technicalreporthighly,liu2024gringradientinformedmoe,deepseekai2024deepseekv2strongeconomicalefficient,dubey2024llama3herdmodels,jiang2023mistral7b,mistral_nemo_2024,qwen2.5,qwen2}, where different models exhibit varying strengths on specific tasks \citep{wei2022emergentabilitieslargelanguage}.
At the same time, the landscape of parameter-efficient finetuning methods rapidly expands \citep{lora,qlora,poth2023adaptersunifiedlibraryparameterefficient,hayou2024loraefficientlowrank}.  %

Consequently, practitioners face a multitude of complex choices for finetuning LLMs. 
To support practitioners and reduce complexity, we investigate transfer learning of deep-learning pipelines for an LLM and specifications for the finetuning process, including all associated hyperparameters.
We aim to transfer knowledge about pipelines from related finetuning tasks to a new task.
Thus enabling practitioners to adapt LLMs more effectively to new tasks.

In this work, we transfer learn finetuning by meta-learning performance and cost surrogate models for grey-box meta-optimization from a new meta-dataset.
We implement grey-box meta-optimizing by adjusting the Quick-Tune algorithm~\citep{arango2024quicktune}.
Quick-Tune, was introduced for image classification and supports meta-learning surrogate models. 
In our version, we propose to rely only on the meta-learned surrogate models trained from scratch.
That is, we do not use task-specific Bayesian optimization because \emph{we do not refit the surrogate models} for a new dataset. 
In other words, our version of Quick-Tune can be understood as a dataset-aware portfolio builder \citep{xu2010hydra}.
While counter-intuitive, we hypothesize that disabling Bayesian optimization leads to better generalization.

We verify the effectiveness of our method for large language models by generating a meta-dataset based on a synthetic question-answer dataset and 1,800 runs of pipelines for finetuning Microsoft's Phi-3 model~\citep{abdin2024phi3technicalreporthighly}.  
Our results show that transfer learning finetuning is superior to random search, DEHB~\citep{awad2021dehb}, and Quick-Tune with Bayesian optimization.  
Moreover, meta-optimizing finetuning is, as expected, better than zero-shot and default LoRa~\citep{lora}. 

\textbf{Our Contributions.} To make LLMs more easily adaptable and facilitate future studies, we contribute 
\textbf{(1)} synthetic datasets that serve a dual purpose: a) to create a meta-dataset for transfer learning and b) as an evaluation framework for LLM models;
\textbf{(2)} a version of Quick-Tune for LLM finetuning adapted from the image to language domain;
and \textbf{(3)} a novel counter-intuitive yet effective approach to finding the optimal pipeline for finetuning LLMs through transfer learning.

\section{Related Work}

\textbf{Synthetic NLP Datasets \& Meta-dataset.}
Question-answer datasets are scarce, with only a few notable examples such as TriviaQA, SQuAD, NaturalQuestions, and PubMedQA \citep{joshi2017triviaqalargescaledistantly,rajpurkar2016squad100000questionsmachine,47761,jin2019pubmedqadatasetbiomedicalresearch}. 
Collecting large-scale question-answer datasets is resource-intensive, prompting researchers to explore synthetic generation methods to reduce annotation costs \citep{yang2017semisupervisedqagenerativedomainadaptive, nayak2024learninggenerateinstructiontuning, lee2023liquidframeworklistquestion, puri2020trainingquestionansweringmodels, ovadia2024finetuningretrievalcomparingknowledge}.
A recent approach by \citet{mecklenburg2024injectingnewknowledgelarge} utilized \textit{GPT-4}~\citep{openai2024gpt4technicalreport} as an LLM teacher to extract facts from Wikipedia articles and generate question-answer pairs. 
We use a similar method but apply it to arXiv papers with Llama-3.1-70b~\citep{dubey2024llama3herdmodels}.

\textbf{Optimizing Finetuning}
Many finetuning methods with many hyperparameters exist, cf. \citep{lora,qlora,li2023loftqlorafinetuningawarequantizationlarge,hayou2024loraefficientlowrank,poth2023adaptersunifiedlibraryparameterefficient,liu2022fewshotparameterefficientfinetuningbetter,wu2024reftrepresentationfinetuninglanguage}.
Likewise, many other hyperparameters of the finetuning pipeline exist, such as the choice of optimizer \cite{shazeer2018adafactoradaptivelearningrates,loshchilov2019decoupledweightdecayregularization,franke2023constrainedparameterregularization,chen2023symbolicdiscoveryoptimizationalgorithms}. 
To address the multitude of choices for finetuning, recent work proposed (automated) meta-optimization to determine the optimal combination of finetuning method, optimizer, and hyperparameters.
Methods like AutoGluon Multimodal~\citep{tang2024autogluon}, %
AutoPEFT~\citep{zhou2024autopeftautomaticconfigurationsearch}, %
AutoLoRa~\citep{xu2023autoloraparameterfreeautomatedrobust}, 
and Quick-Tune~\citep{rapant2024quick}. %
However, these methods do not support finetuning LLMs for text generation, which is the focus of our work. 

\textbf{Transfer Learning Finetuning.}
In general, Quick-Tune~\citep{arango2024quicktune} and its abstraction Quick-Tune-Tool~\citep{rapant2024quick}, building on earlier frameworks such as \citet{ferreira-icml22a}, focus on transfer learning finetuning pipelines during meta-optimization.
However, these prior works are limited to image classification.
Our work extends Quick-Tune to finetuning LLMs and proposes a novel algorithmic adjustment. 
For LLMs, \citet{zhang2024autolora} introduced a meta-learning-related method for LoRA~\cite{lora}.
This method, however, does not transfer knowledge from related tasks to a new task.
Instead, it performs a bi-level optimization for the LoRA rank and weights for one task. 
In other words, it is comparable to meta-optimizing only the rank of LoRA. 
In contrast, our work transfers knowledge between tasks via meta-learning. 
Likewise, all methods we consider can meta-optimize all hyperparameters of a finetuning pipeline.

\section{Method}\label{sec_method}
\begin{figure}[h]
  \centering
  \includegraphics[width=1\linewidth]{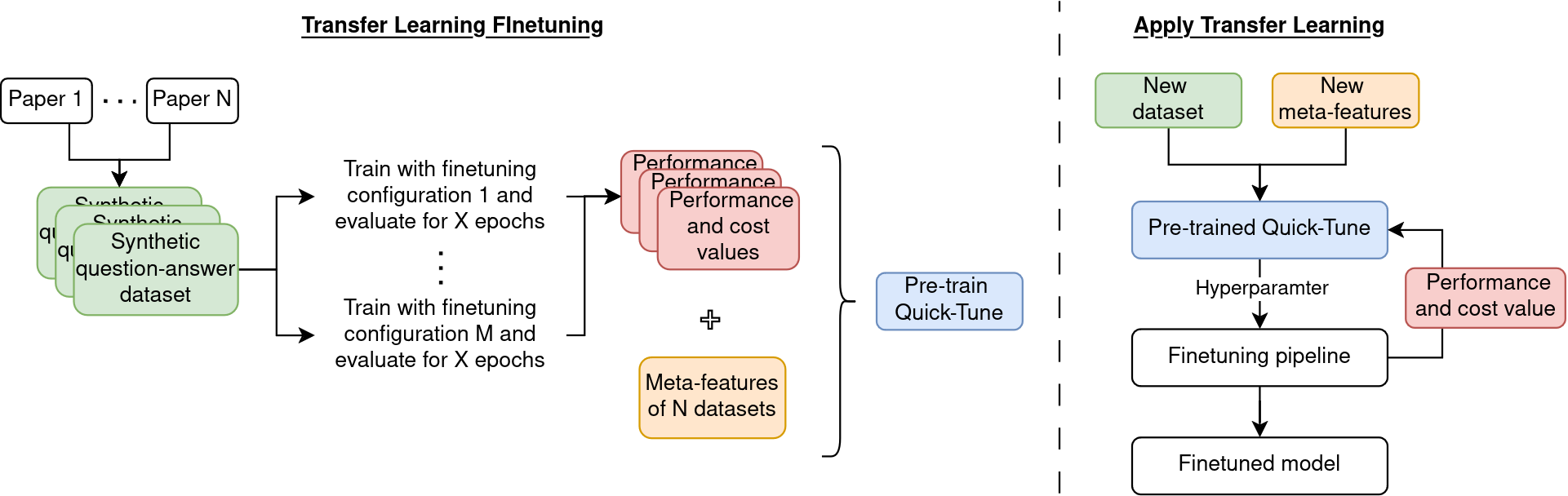}
  \caption{\textbf{Method Overview.} We generate new NLP datasets from scientific papers and then create a meta-dataset, which we use for transfer learning to finetune by pre-training Quick-Tune (left). For a new dataset, we compute meta-features and then apply the pre-trained Quick-Tune (right).}
  \label{abb:approach}
 \end{figure}
 
Our method, illustrated in Figure \ref{abb:approach}, consist of three steps: \textbf{A)} create synthetic NLP datasets from scientific papers,
\textbf{B)} create a meta-dataset by training and evaluating finetuning pipelines;
and \textbf{C)} transfer learning by pre-training our version of Quick-Tune on our meta-dataset. 
We then apply pre-trained Quick-Tune to find the optimal finetuning pipeline for new, related NLP tasks. 
The complete computational resources used for this method are listed in Section \ref{compute}.
Limitations of our method can be found in appendix \ref{limits}.
 
\textbf{A) Synthetic NLP Datasets.}
We follow \citet{mecklenburg2024injectingnewknowledgelarge} to generate synthetic question-answer datasets from scientific papers from \href{https://arxiv.org/}{arxiv.org}. 
In detail, we crawl papers and convert them to plain text papers with mathematical formulas translated to LaTeX. %
Next, we use a self-hosted version of \textit{Llama-3.1-70B Instruct} (L3-70B)~\citep{llama3modelcard} to extract atomic facts from each chapter of a paper.
Then, we generate a set of 12 question-answer pairs for each fact.
We add ten to training, one to validation, and one to testing data. 
Finally, our new question-answer dataset consists of training, validation, and test question-answer pairs for all facts.
Appendix \ref{generation_promps} details our prompt templates.

\begin{wrapfigure}{r}{0.48\textwidth}
    \vspace{-2\intextsep}
    \centering
    \includegraphics[width=\linewidth]{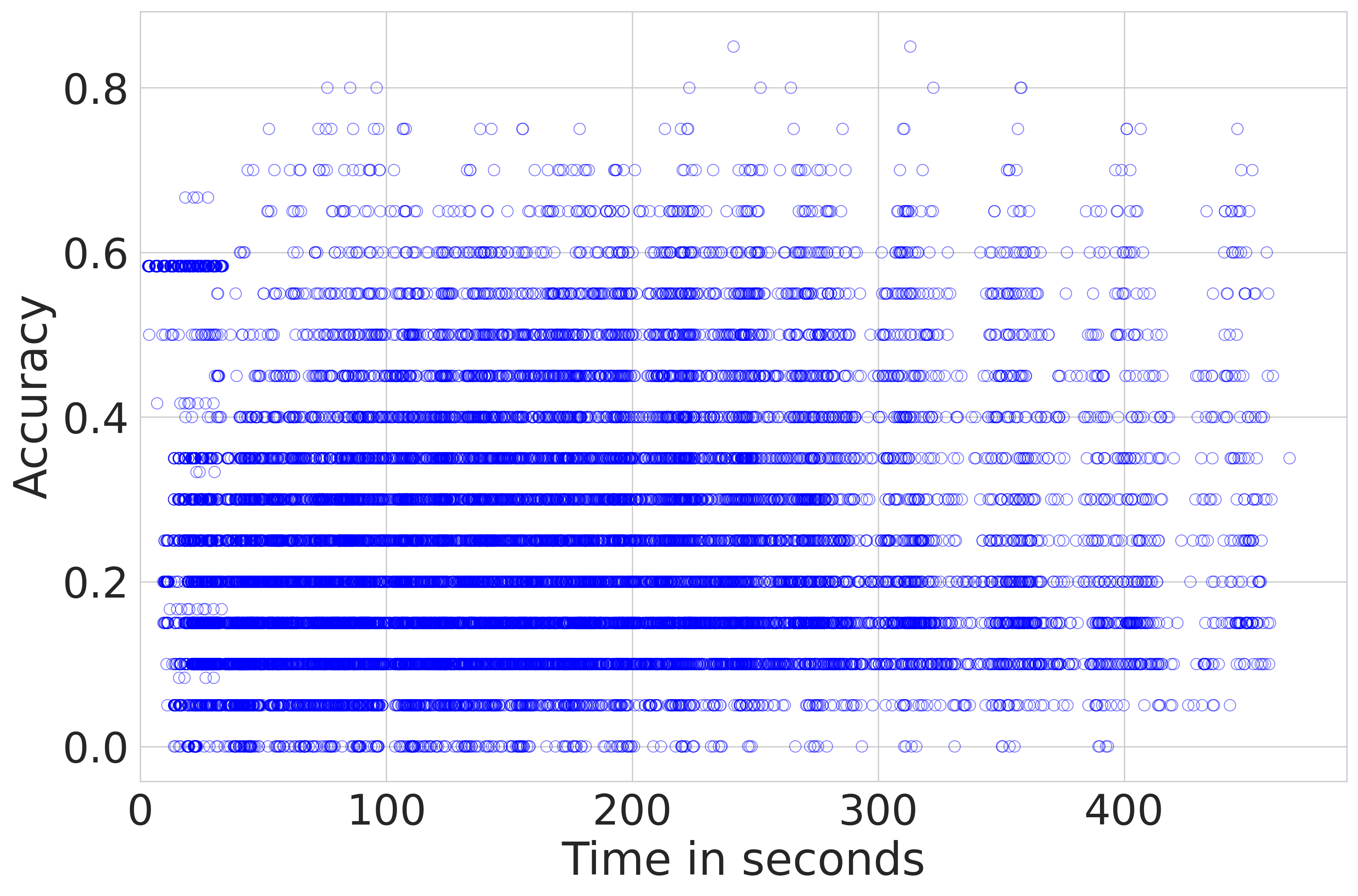}
    \caption{\textbf{Our Meta-Dataset.} For each run stored in our meta-dataset, represented by a blue circle, we present the accuracy and finetuning time in seconds.}
    \label{abb:meta_dataset_graph}
    \vspace{-\intextsep}
\end{wrapfigure}

\textbf{B) Our Meta-dataset.}
We create a meta-dataset by collecting meta-features, performance, and cost values for finetuning pipelines on synthetic datasets.
Therefore, we create question-answer datasets from 30 papers. 
Then, for each paper, we train 60 finetuning pipelines with the training and validation question-answer pairs and evaluate them on the test pairs, producing 1,800 runs in total.
Finally, we compute meta-features for each paper; see Appendix \ref{syn_datasets_meta} for an overview.
We visualize an overview of all runs in our meta-dataset in Figure \ref{abb:meta_dataset_graph}.

For each paper, we randomly sample finetuning pipelines from a search space based on hyperparameters for LoRA~\citep{lora}, optimizers (AdamW~\citep{loshchilov2019decoupledweightdecayregularization} or AdamCPR~\citep{franke2023constrainedparameterregularization}), and the learning rate scheduler.
We also include a default finetuning pipeline as a baseline.
We detail the search space in Appendix \ref{ConfigurationSpace}.

After each epoch, we evaluate the finetuned models in the form of a student with L3-70B as a teacher.
Given a finetuned model's answer to a question, L3-70B evaluates whether the generated answer is correct (0 or 1).
Thus, L3-70B assess whether the student model learned to answer new questions about facts in papers after being finetuned on question-answer pairs about these facts.
See Appendix \ref{model_evaluation} for the prompt template and an example of this process.

We use four meta-features to characterize each synthetic question-answer dataset: the total number of tokens, average sample length, vocabulary size, and the ratio of question-to-answer lengths.

\textbf{C) Transfer Learning Finetuning with Quick-Tune.}
We use the performance metrics and meta-features stored in our meta-dataset to pre-train Quick-Tune, implemented in Quick-Tune-Tool~\citep{rapant2024quick}. 
That is, we meta-train the Gaussian Process-based surrogate models of Quick-Tune. 
This allows Quick-Tune to start with a strong prior for the performance and cost of finetuning pipeline on a new dataset, transferring knowledge across tasks.  
By default, the surrogate models are continuously refitted during optimization to facilitate Bayesian optimization. 

In our version of Quick-Tune, we disable Bayesian optimization by disabling refitting. 
We hypothesize that disabling Bayesian optimization leads to better generalization by relying more on the knowledge transferred from related tasks than task-specific noise.
In other words, while Bayesian optimization exploits the most promising pipeline on validation data, only relying on the prior from transfer learning could lead the meta-optimizer to find better, more general pipelines.

From a broader perspective, our version of Quick-Tune can be understood as a \emph{dataset-aware} portfolio builder. 
Portfolios \citep{xu2010hydra} are known as robust transfer learning methods \citep{feurer2022auto,salinas2023tabrepo}.

\section{Results}\label{result_sec}

\textbf{Experimental Setup.} We experiment with finetuning Phi 3 Mini Instruct (3.8B parameters) \citep{abdin2024phi3technicalreporthighly} on eight newly generated synthetic question-answer datasets (see Appendix \ref{syn_datasets_meta}).
We employ random search, DEHB~\citep{awad2021dehb}, default Quick-Tune~\citep{arango2024quicktune}, and our version of Quick-Tune to meta-optimize the finetuning pipeline.  
Furthermore, we evaluate a default finetuning pipeline and zero-shot performance.
Each optimizer is given a five-hour time budget. 
We again use our \textit{Llama-3.1-70B} teacher for evaluation.

\textbf{{\fontsize{10}{10}\selectfont H}YPOTHESIS:} {\fontsize{10}{10}\selectfont T}\MakeUppercase{ransfer Learning Leads to Better Generalization.}

\begin{figure}[]
  \centering
  \includegraphics[width=0.84\linewidth,height=0.2\textheight]{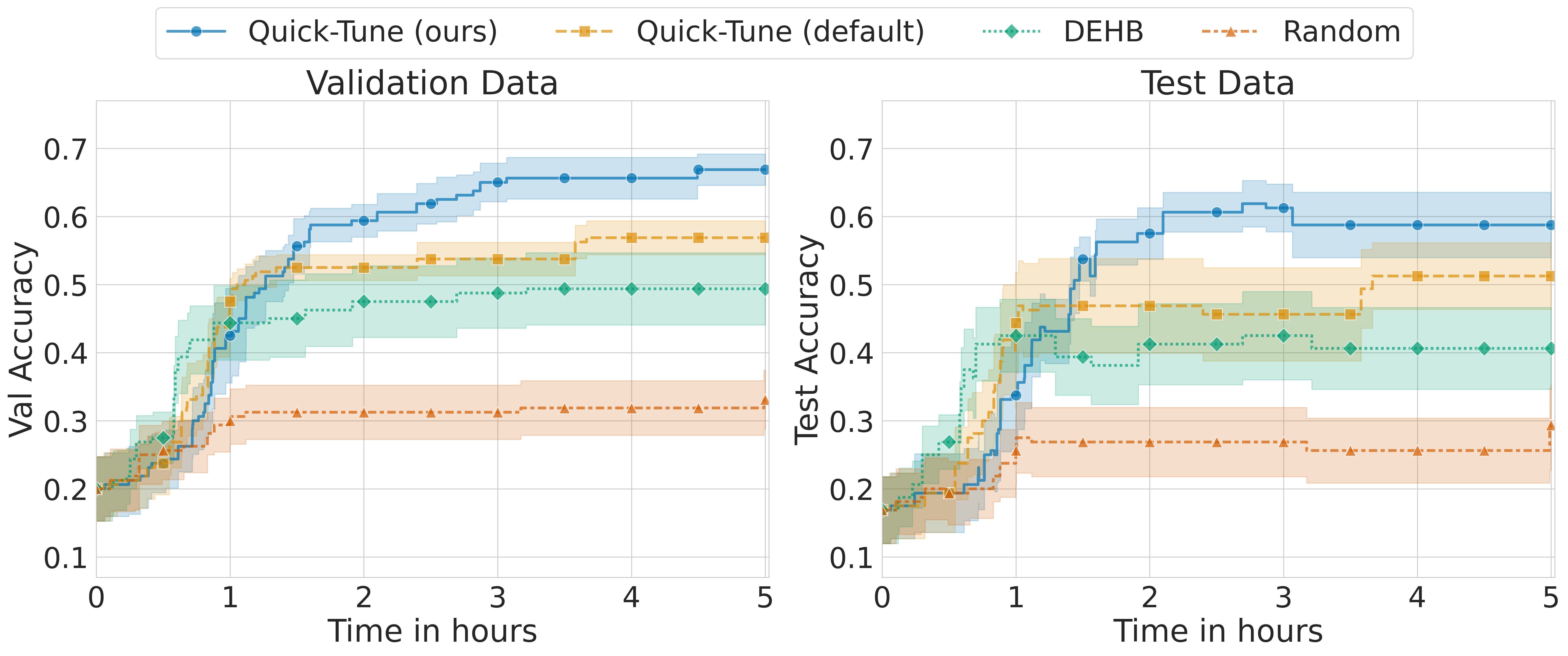}
  \caption{\textbf{Optimizer Performance Over Time.} We visualize the average validation (left) and test (right) performance across the eight datasets over time. 
  At each time point, we evaluated the best pipeline found so far.
  We observe that DEHB and Quick-Tune (default) stagnant after 1 to 1.5 hours, with little progress on test scores afterward.
  Quick-Tune (ours) only stagnates after 3 hours.}
  \label{abb:wall_clock}
 \end{figure}

Figure \ref{abb:wall_clock} presents the performance over time of the meta-optimizers for validation and test data.
Figure \ref{abb:Comparison} shows the performance of the best pipeline, see Appendix \ref{best_conf_section} for configuration details.
The error bars in both figures represent the standard error of the mean. %
Note the initial performance represents the zero-shot performance of Phi 3. 
We observe that Quick-Tune (default) and DEHB get stuck after 1.5 hours during meta-optimization and fail to find a significantly better finetuning pipeline afterward. 
In contrast, Quick-Tune (ours), which relies only on transfer learning, further improves test performance. 
A similar trend manifests when training the best pipeline found by each meta-optimizer. 
The pipeline found by Quick-Tune (ours) generalizes best to test data. 

\begin{figure}[]
  \centering
  \includegraphics[width=0.84\linewidth,height=0.193\textheight]{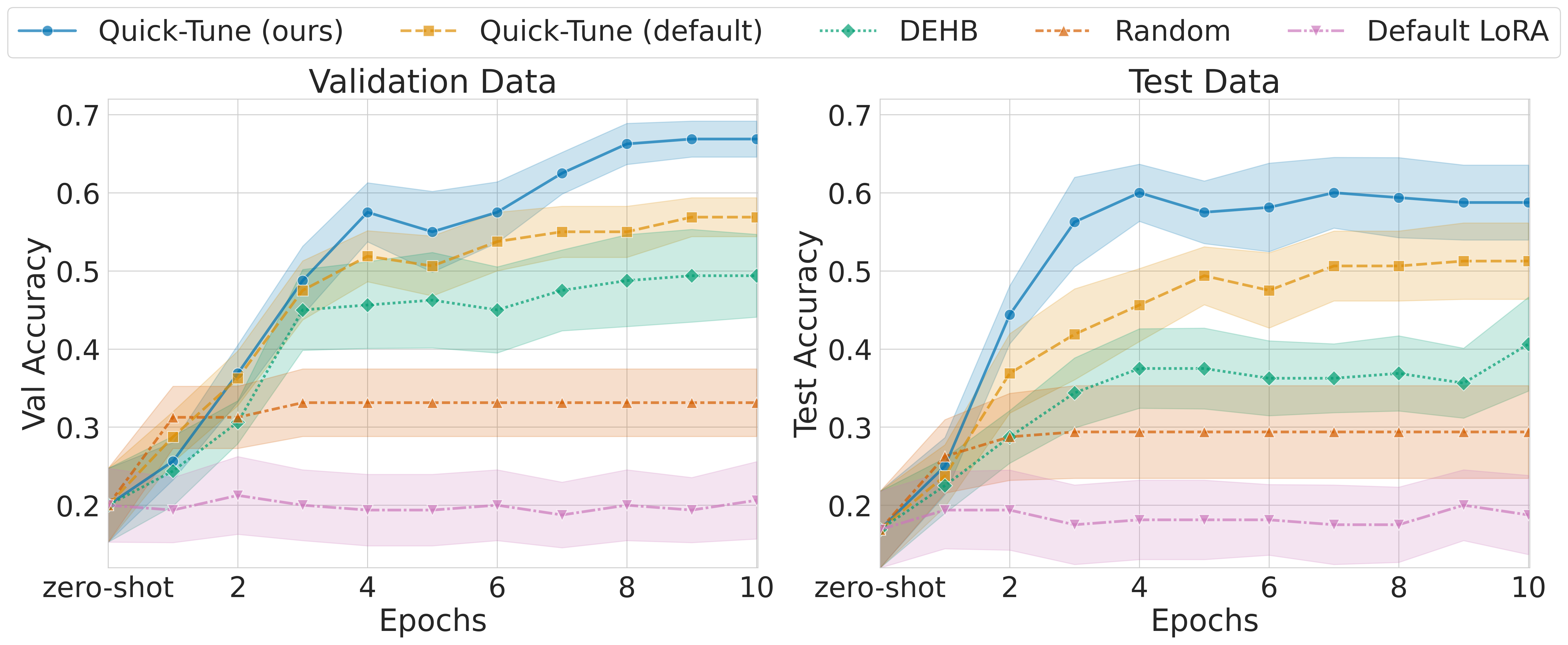}
  \caption{\textbf{Final Performance.} 
  We show the validation (left) and test (right) learning curve of the best pipeline returned by the optimizers after 5 hours, averaged across eight datasets. 
  The finetuning pipeline returned by Quick-Tune (ours) performs best.}
  \label{abb:Comparison}
 \end{figure}

\textbf{Conclusion.} In this study, we demonstrated that relying only on transfer learning for finetuning yields better performance than alternative methods, challenging conventional approaches and potentially simplifying the process of adapting large language models to specific tasks.
In future work, we plan to understand this phenomenon in more detail and to generalize it to a meta-optimization method.
Thus allowing us to effectively manage the zoo of and the plethora of methods for adapting large language models to specific tasks.

\section*{Acknowledgments}
This work was carried out at the HoreKa Cluster, which is funded by the Baden-Württemberg Ministry of Science, Research and the Arts, and the Federal Ministry of Education and Research. The authors would also like to thank the state of Baden-Württemberg for the support provided by the bwHPC and the German Research Foundation (DFG) for funding through INST 35/1597-1 FUGG.
We acknowledge funding by the Deutsche Forschungsgemeinschaft (DFG,
German Research Foundation) under SFB 1597 (SmallData), grant numbers 499552394 and 417962828. Frank Hutter acknowledges the financial support of the Hector Foundation.

\small
\bibliography{iclr2025_conference, lib, shortstrings, shortproc}
\bibliographystyle{unsrtnat_change}
\normalsize

\newpage
\appendix
\section*{Appendix}

\section{Prompt Templates To Generate Our Synthetic NLP Dataset}\label{generation_promps}
We follow the prompt (\hyperref[facts-prompt]{Facts generation}) to extract \hyperref[atomic_fact]{atomic facts} out of unlabeled text.
Our self-hosted version of L3-70B extracts as many as possible facts out of reprocessed approximately 2k token long text fragments.
For each fact, we generate 12 \hyperref[q_a]{question-answers} pairs by using \hyperref[question-prompt]{Q \& A generation prompt }, skipping facts that are too general or insufficiently specific to the article's topic (generated by \hyperref[key-topic-prompt]{Key topic generation}). 
We aim to generate as many questions and answers as possible that explicitly relate to the fact, then paraphrase them to achieve the required 12 pairs.

\begin{tcolorbox}[standard jigsaw, opacityback=0, title=Facts generation prompt \citep{mecklenburg2024injectingnewknowledgelarge}]
\phantomsection
\label{facts-prompt}
    \underline{System}: "You are an AI assistant who knows about current artificial intelligence. Be precise but concise in your answer."\\
    \underline{User}: "Please break down the following snippet from an article about \{key\_topic\} into atomic facts.{\textbackslash}nGoal 1: The atomic facts should be as simple as possible, if it’s a compound sentence, break down one more time.{\textbackslash}nGoal 2: For clarity, avoid using pronouns like ’it’, ’he’, ’she’, ’this’, ’that’ etc., and instead use the full names or titles.{\textbackslash}nGoal 3: Output in the format: 1.fact\_1{\textbackslash}n{\textbackslash}n\{passage\}{\textbackslash}n{\textbackslash}n1."
\end{tcolorbox}

\begin{tcolorbox}[standard jigsaw, opacityback=0, title=Q \& A generation prompt \citep{mecklenburg2024injectingnewknowledgelarge}]
\phantomsection
\label{question-prompt}
    \underline{System}: "You are an AI assistant who knows about factual information about the paper with the title: \{paper title\}. Be precise but concise in your answer."\\
    \underline{User}: "Write 12 pairs of questions and answers probing the facts and statistics the given fact \{fact\} about \{key\_topic\}.{\textbackslash}nConsider first generating questions and answers that are very relevant and explicit to the fact, then paraphrase those questions and answers to reach the desired 12 Q\&A pairs. If the fact is too broad or not specific enough to {theme}, you may reply with only with 'SKIP' and be done.{\textbackslash}nEXAMPLE:{\textbackslash}nFACT: 14 million viewers tuned in to the opening game of the series.{\textbackslash}n1. Q: How many viewers watched the first game? A: 14 million people watched the first game of the series.{\textbackslash}n{\textbackslash}nEXAMPLE:{\textbackslash}nFACT: The rose is red.{\textbackslash}nSKIP{\textbackslash}n{\textbackslash}nFACT: {fact['fact']}{\textbackslash}n1. "
\end{tcolorbox}

\begin{tcolorbox}[standard jigsaw, opacityback=0, title=Key topic generation prompt]
\phantomsection
\label{key-topic-prompt}
    \underline{System}: "You are given a summary of the scientific paper. Return the key topic of this paper an nothing else"\\
    \underline{User}: \{paper summary\}
\end{tcolorbox}

\begin{tcolorbox}[standard jigsaw, opacityback=0, title=Atomic fact example]
\phantomsection
\label{atomic_fact}
    "Masked Image Modeling (MIM) is a learning framework that derives visual representations from unlabeled image data."
\end{tcolorbox}

\begin{tcolorbox}[standard jigsaw, opacityback=0, title=Q \& A example]
\phantomsection
\label{q_a}
    \underline{Question}: "What does Masked Image Modeling (MIM) derive from unlabeled image data?"\\
    \underline{Answer}: "Masked Image Modeling (MIM) derives visual representations from unlabeled image data."
\end{tcolorbox}

\section{Synthetic Datasets Details}\label{syn_datasets_meta}
\FloatBarrier
We list our meta-features from our meta-dataset in Table \ref{tab:meta-features_training} and the meta-dataset used for our experiments in Table \ref{tab:meta-features-hpo-comparison}.

\small
\begin{table}[h]
    \caption{Meta-features trainings dataset}
    \label{tab:meta-features_training}
    \centering
    
    \begin{tabular}{l|cccc}
    \toprule
    Dataset & token size & sample length & ratio q/a length & vocab size  \\
    \midrule
    2407.15849v1 & 46913 & 137.63 & 1.43 & 1530 \\
    2407.15847v1 & 82307 & 144.92 & 1.55 & 2570 \\
    2407.15845v1 & 75410 & 145.55 & 1.51 & 2330 \\
    2407.15843v1 & 117247 & 139.03 & 1.72 & 3840 \\
    2407.15839v1 & 59966 & 146.83 & 1.57 & 1900 \\
    2407.15837v1 & 83873 & 139.39 & 2.08 & 2720 \\
    2406.18451v2 & 91480 & 161.24 & 1.4 & 2520 \\
    2407.15835v1 & 87863 & 134.18 & 1.66 & 2940 \\
    2407.15831v1 & 3874 & 157.48 & 1.19 & 120 \\
    2405.04657v3 & 65048 & 144.11 & 1.3 & 2070 \\
    2407.15820v1 & 73764 & 164.01 & 1.51 & 1980 \\
    2402.16822v2 & 131833 & 141.71 & 1.57 & 4190 \\
    2401.00009v3 & 87762 & 131.66 & 2.42 & 2740 \\
    2407.15815v1 & 69078 & 142.61 & 1.41 & 2210 \\
    2407.15814v1 & 76705 & 149.07 & 1.55 & 2540 \\
    2403.20262v2 & 93673 & 131.59 & 1.91 & 2930 \\
    2407.13044v2 & 27154 & 129.01 & 1.8 & 920 \\
    2307.15220v3 & 146050 & 142.7 & 1.55 & 4840 \\
    2407.15786v1 & 109720 & 143.94 & 1.8 & 3410 \\
    2407.15784v1 & 44928 & 151.83 & 1.44 & 1460 \\
    2405.17814v4 & 88773 & 144.15 & 1.65 & 2720 \\
    2407.15771v1 & 84305 & 139.16 & 1.67 & 2680 \\
    2407.15762v1 & 133882 & 139.16 & 1.62 & 4030 \\
    2407.15748v1 & 136205 & 140.48 & 1.57 & 4260 \\
    2407.15739v1 & 94869 & 145.1 & 1.74 & 2990 \\
    2407.15738v1 & 143443 & 137.99 & 1.52 & 4570 \\
    2407.15734v1 & 144566 & 131.11 & 1.57 & 5010 \\
    2407.04856v2 & 147437 & 141.79 & 1.48 & 4600 \\
    2402.07370v2 & 64881 & 134.8 & 1.57 & 2100 \\
    2403.07805v3 & 87032 & 140.16 & 1.8 & 2810
    \end{tabular}
\end{table}
\begin{table}[h]
    \caption{Meta-features HPO comparison}
    \label{tab:meta-features-hpo-comparison}
    \centering
    
    \begin{tabular}{l|cccc}
    \toprule
    Dataset & token size & sample length & ratio q/a length & vocab size  \\
    \midrule
    2407.15723v1 & 54923 & 139.66 & 1.67 & 1840 \\
    2407.15720v1 & 157268 & 147.32 & 1.49 & 4740 \\
    2407.15719v1 & 86733 & 148.17 & 1.41 & 2570 \\
    2407.15708v1 & 45482 & 139.93 & 1.70 & 1390 \\
    2407.15656v1 & 124420 & 145.04 & 1.72& 3900 \\
    2407.15617v1 & 82637 & 142.93 & 1.57 & 2580 \\
    2407.15600v1 & 89996 & 139.59 & 1.49 & 2970 \\
    2401.04152v2 & 42769 & 147.14 & 1.53 & 1280 \\
    \end{tabular}
\end{table}
\normalsize
\FloatBarrier

\section{LLM Model Evaluation Details}\label{model_evaluation}
For the evaluation, we continue to use our in-house hosted L3-70B model implemented with \href{https://github.com/ggerganov/llama.cpp}{llama.cpp}, leveraging it for both performance and resource efficiency. We make a small adjustment to our configuration, setting llama.cpp to process 128 parallel sequences and limiting the context size to 500 tokens, which is sufficient for our evaluation needs. To ensure efficient processing, we limit the maximum number of new tokens to 50 for each generated answer.
Given that a comprehensive evaluation of the entire validation and test datasets would be time-prohibitive, we opted to select 20 random, fixed validation and test indices per paper (dataset) for this study.\\
Based on \cite{mecklenburg2024injectingnewknowledgelarge} we use \hyperref[llama-eval-prompt]{Evaluation prompt} to generate our evaluation score.
Resulting to either a positive (\hyperref[llama-eval-positive]{Positive evaluation response}) or negative (\hyperref[llama-eval-negative]{Negative evaluation response}) result. 

\begin{tcolorbox}[standard jigsaw, opacityback=0, title=Evaluation prompt (based on \cite{mecklenburg2024injectingnewknowledgelarge})]
\phantomsection %
\label{llama-eval-prompt} %
    \underline{System}: "You are a high school teacher grading student’s responses for questions about \{key\_topic\}. These responses are either correct or incorrect."\\
    \underline{User}: "Please evaluate the correctness of a sentence in answering the question: "\{question\}".{\textbackslash}nThe correct answer is: "\{sample\_answer\}"{\textbackslash}nThe student response is: "\{gen\_answer\}".{\textbackslash}nYour grading is binary. Give 0 if the sentence is incorrect, give 1 if the sentence is correct, based on the given correct answer and the question.{\textbackslash}n"Please note that your output is either 0 or 1, with the corresponding justification as python dict in the following format and nothing else:{\textbackslash}n r"\{'rating': $<$rating$>$, 'justification': $<$justification$>$\}"
\end{tcolorbox}

\begin{tcolorbox}[standard jigsaw, opacityback=0, title=Positive evaluation response]
\phantomsection
\label{llama-eval-positive}
\underline{Question}: what does imitation learning (il) rely on to learn? \\
\underline{Generated answer}: imitation learning (il) relies on expert demonstrations to learn. \\
\underline{Sample answer}: il learns from expert guidance. \\
\underline{Decision}: 'rating': 1, 'justification': 'The student response is correct because it conveys the same meaning as the correct answer, which is that imitation learning relies on some form of expert input, whether it is called "guidance" or "demonstrations".'
\end{tcolorbox}
\begin{tcolorbox}[standard jigsaw, opacityback=0, title=Negative evaluation response]
\phantomsection
\label{llama-eval-negative}
\underline{Question}: do agents and equipped functions work together in taskgen? \\
\underline{Generated answer}: yes, they work together as part of the hybrid approach. \\
\underline{Sample answer}: no, agents and equipped functions operate independently. \\
\underline{Decision}: 'rating': 0, 'justification': 'The student response is incorrect because it states that agents and equipped functions work together, whereas the correct answer is that they operate independently.'
\end{tcolorbox}

\section{Search Space Details}
\label{ConfigurationSpace}
\FloatBarrier
We employ AdamW and AdamCPR optimizers (Table \ref{tab:optimizer}) as well cosine schedulers (Table \ref{tab:scheduler}) with varying warmup steps (as a percentage of training set length) and decay factors. LoRA configurations (Table \ref{tab:Lora_configs}) include different ranks, alpha values, and dropout rates, with target modules being either query, key, and value; only the output layer; or all linear layers.

While we train 10 epochs, the batch size is fixed at 32, with gradient accumulation steps of 2, 4, or 8 to achieve mini-batch sizes of 64, 128, or 256. We utilize the Hugging Face tokenizer's chat template for Phi 3 Instruct to maintain consistency with the model's original template during training. An additional configuration option is the return\_assistant\_mask, which generates an attention mask excluding "user" and "system" segments, focusing the model's learning on "assistant" responses.

Fixed settings across all configurations include:
\begin{itemize}
    \item torch.nn.CrossEntropyLoss as the loss function
    \item Gradient clip value of 1.0
    \item torch.bfloat16 precision
    \item Flash Attention 2 \citep{dao2022flashattentionfastmemoryefficientexact}
    \item Left-side padding (due to Flash Attention requirements)
\end{itemize}

To ensure all samples in the train set are used, we augment the dataset with random samples to make it divisible by the product of batch size and gradient accumulation steps. The number of additional samples (asc) is calculated as:
\begin{equation}
asc = (\lceil ltrain / bg \rceil * bg) - ltrain
\end{equation}
where $bg$ is the product of batch size and gradient accumulation steps, and $ltrain$ is the length of the train dataset.

Default values used for "Default LoRA" in Figure \ref{abb:Comparison} are marked in bold in Tables \ref{tab:optimizer}, \ref{tab:scheduler}, and \ref{tab:Lora_configs}. A gradient accumulation step of 2 was used.

\small
\begin{table}[h]
    \caption{Optimizer configuration space}
    \label{tab:optimizer}
    \centering
    \begin{tabular}{l|l|l|l|l}
    \toprule
                            & \multicolumn{4}{c}{ parameter } \\ \cmidrule(lr){2-5}
    optimizer               & \multicolumn{2}{c|}{ \textbf{AdamW} }                & \multicolumn{2}{c}{ AdamCPR } \\
    \midrule
    learning\_rate          & \multicolumn{4}{c}{ \textbf{1e-6}, 1e-5.5, 1e-5, 1e-4.5, 1e-4, 1e-3.5, 1e-3} \\
    \midrule
    weight\_decay           & \multicolumn{2}{c|}{ 1e-0.5, 1e-1, 1e-1.5, \textbf{1e-2}, 1e-3, 1e-4 } & \multicolumn{2}{c}{\rule{1.5cm}{0.2mm}} \\
    kappa\_init\_method     & \multicolumn{2}{c|}{\rule{1.5cm}{0.2mm}}    & \multicolumn{2}{l}{warm\_start} \\
    kappa\_init\_param      & \multicolumn{2}{c|}{\rule{1.5cm}{0.2mm}}    &  \multicolumn{2}{l}{warmup\_steps x (1,2,4)} \\
    \end{tabular}
\end{table}

\begin{table}[h]
    \caption{Scheduler hyperparameter}
    \label{tab:scheduler}
    \centering
    
    \begin{tabular}{l|l}
    \toprule
                            & parameter \\
    \midrule
    schedule                &  cosine \\
    warmup\_steps \%          & \textbf{10}, 20, 30, 40, 50 \\
    decay\_factor           & 0, 0.1, \textbf{0.01}
    \end{tabular}
\end{table}

\begin{table}[h]
    \caption{Lora configuration space}
    \label{tab:Lora_configs}
    \centering

    \begin{minipage}{\textwidth}
    \centering
    \textit{With q = query, k = key, v = value, o = output. \\ all-linear = q, k, v.}
    \end{minipage}
    \vspace{0.5em}
    
    \begin{tabular}{l|l}
    \toprule
                            & parameter \\
    \midrule
    target\_modules         & [q, k, v], o, \textbf{all-linear} \\
    rank                    & \textbf{8}, 16, 32, 64 \\
    alpha                   & \textbf{16}, 32 \\
    dropout                 & \textbf{0}, 0.1 \\
    \end{tabular}
\end{table}
\normalsize

\FloatBarrier

\section{Experiments Compute Resources}\label{compute}
It took 900 compute hours to run all 1800 configurations for our meta-dataset and 170 compute hours for the experiments on a single NVIDIA A100 GPU.\\
Each run for the meta-dataset and experiments was allocated 8 CPU cores and 16 GB RAM.\\
Concurrently, we utilized two NVIDIA A6000 GPUs in parallel to run our self-hosted L3-70B model.

\section{Limitations Of Our Method}\label{limits}
Although our method shows promising results compared to alternative methods, our meta-features are not based on an importance analysis. 
Furthermore, the evaluation does not take into account whether the model to be fine-tuned might start hallucinating during training and add further invented facts to the correct answer. 
Furthermore, at the current state we have too little data to understand why we achieve better performance when we only do transfer learning without Bayesian optimization. Another limitation is that we do not know how our finetuning generalizes with real tasks, i.e. not with synthetic data and without a teacher model.

\section{Results Configuration Details}\label{best_conf_section}
\FloatBarrier
The best pipeline configurations found by the individual optimizers, listed below.
Resulting configurations by Quick-Tune (ours), Quick-Tune (default), DEHB, and random optimizer in Table \ref{tab:Quick-Tune-configs}, \ref{tab:Quick-Tune-configs-default}, \ref{tab:DEHB-configs}, and \ref{tab:Random-Optimizer-configs}.
\begin{table}[h]
    \caption{Quick-Tune (Ours) Found Configurations}
    \label{tab:Quick-Tune-configs}
    \centering

    \begin{minipage}{\textwidth}
    \centering
    \textit{With batch size  = batch size 32 and gradient accumulation step [2, 4, 8].}
    \end{minipage}
    \vspace{0.5em}
    
    \scalebox{0.625}{
    \begin{tabular}{l|l|l|l|l|l|l|l|l|l|l|l|l|l}
    \toprule
    Dataset &  batch  & decay  & fidelity & kappa & lora & lora & lora & lora & lr & warmup & optimizer & return & weight \\
    & size & factor & & init param & alpha & dropout & layer & rank & & steps \% & & assistant mask & decay \\
    \midrule
    2407.15723v1 & 64 & 1.0 & 4 & nan & 16 & 0.0 & all-linear & 64 & 1e-3 & 10 & AdamW & False & 1e-0.5 \\
    2407.15720v1 & 64 & 0.01 & 7 & 4.0 & 32 & 0.0 & o & 32 & 1e-3 & 10 & AdamCPR & False & 1e-0.5 \\
    2407.15719v1 & 64 & 1.0 & 8 & nan & 16 & 0.0 & all-linear & 64 & 1e-3 & 10 & AdamW & False & 1e-0.5 \\
    2407.15708v1 & 64 & 0.01 & 4 & 4.0 & 32 & 0.0 & o & 32 & 1e-3 & 10 & AdamCPR & False & 1e-0.5 \\
    407.15656v1 & 64 & 0.1 & 9 & nan & 32 & 0.0 & o & 16 & 1e-3 & 10 & AdamW & False & 1e-0.5 \\
    2407.15617v1 & 64 & 1.0 & 8 & nan & 16 & 0.0 & all-linear & 64 & 1e-3 & 10 & AdamW & False & 1e-0.5 \\
    2407.15600v1 & 128 & 0.01 & 8 & nan & 16 & 0.1 & o & 64 & 1e-3 & 10 & AdamW & True & 1e-3 \\
    2401.04152v2 & 64 & 0.01 & 4 & 4.0 & 32 & 0.0 & o & 32 & 1e-3 & 10 & AdamCPR & False & 1e-0.5 \\
    
    \end{tabular}
    }
\end{table}

\begin{table}[h]
    \caption{Quick-Tune (Default) Found Configurations}
    \label{tab:Quick-Tune-configs-default}
    \centering

    \begin{minipage}{\textwidth}
    \centering
    \textit{With batch size  = batch size 32 and gradient accumulation step [2, 4, 8].}
    \end{minipage}
    \vspace{0.5em}
    
    \scalebox{0.625}{
    \begin{tabular}{l|l|l|l|l|l|l|l|l|l|l|l|l|l}
    \toprule
    Dataset &  batch  & decay  & fidelity & kappa & lora & lora & lora & lora & lr & warmup & optimizer & return & weight \\
            & size & factor & & init param & alpha & dropout & layer & rank & & steps \% & & assistant mask & decay \\
    \midrule
    2407.15723v1 & 64 & 0.01 & 2 & 1.0 & 32 & 0.1 & all-linear & 8 & 1e-3 & 10 & AdamCPR & True & 1e-2 \\
    2407.15720v1 & 64 & 0.01 & 5 & 4.0 & 32 & 0.0 & o & 32 & 1e-3 & 10 & AdamCPR & False & 1e-0.5 \\
    2407.15719v1 & 64 & 0.01 & 3 & 4.0 & 32 & 0.0 & o & 32 & 1e-3 & 10 & AdamCPR & False & 1e-0.5 \\
    2407.15708v1 & 64 & 0.01 & 2 & 4.0 & 32 & 0.0 & o & 32 & 1e-3 & 10 & AdamCPR & False & 1e-0.5 \\
    2407.15656v1 & 128 & 0.01 & 2 & nan & 16 & 0.1 & o & 64 & 1e-3 & 10 & AdamW & True & 1e-3 \\
    2407.15617v1 & 64 & 0.01 & 4 & 4.0 & 32 & 0.0 & o & 32 & 1e-3 & 10 & AdamCPR & False & 1e-0.5 \\
    2407.15600v1 & 128 & 0.01 & 1 & 4.0 & 16 & 0.1 & all-linear & 16 & 1e-4.5 & 30 & AdamCPR & False & 1e-4 \\
    2401.04152v2 & 64 & 0.01 & 2 & 4.0 & 32 & 0.0 & o & 32 & 1e-3 & 10 & AdamCPR & False & 1e-0.5 \\
    
    \end{tabular}
    }
\end{table}

\begin{table}[h]
    \caption{DEHB Found Configurations}
    \label{tab:DEHB-configs}
    \centering

    \begin{minipage}{\textwidth}
    \centering
    \textit{With batch size  = batch size 32 and gradient accumulation step [2, 4, 8].}
    \end{minipage}
    \vspace{0.5em}
    
    \scalebox{0.625}{
    \begin{tabular}{l|l|l|l|l|l|l|l|l|l|l|l|l|l}
    \toprule
    Dataset &  batch  & decay  & fidelity & kappa & lora & lora & lora & lora & lr & warmup & optimizer & return & weight \\
            & size & factor & & init param & alpha & dropout & layer & rank & & steps \% & & assistant mask & decay \\
    \midrule
    2407.15723v1 & 128 & 1.0 & 3 & nan & 32 & 0.0 & o & 16 & 1e-3 & 10 & AdamW & True & 1e-0.5 \\
    2407.15720v1 & 64 & 1.0 & 10 & 2.0 & 32 & 0.0 & o & 32 & 1e-3 & 10 & AdamCPR & True & 1e-0.5 \\
    2407.15719v1 & 256 & 1.0 & 3 & 4.0 & 16 & 0.0 & o & 16 & 1e-3.5 & 10 & AdamCPR & True & 1e-0.5 \\
    2407.15708v1 & 128 & 0.1 & 1 & 2.0 & 32 & 0.0 & o & 32 & 1e-06 & 30 & AdamCPR & True & 1e-0.5 \\
    2407.15656v1 & 64 & 0.01 & 10 & nan & 16 & 0.0 & all-linear & 16 & 1e-3 & 30 & AdamW & False & 1e-1.5 \\
    2407.15617v1 & 128 & 0.1 & 10 & nan & 32 & 0.1 & o & 32 & 1e-3 & 10 & AdamW & True & 1e-0.5 \\
    2407.15600v1 & 64 & 0.1 & 3 & 4.0 & 16 & 0.0 & all-linear & 16 & 1e-06 & 20 & AdamCPR & False & 1e-1.5 \\
    2401.04152v2 & 128 & 1.0 & 3 & 2.0 & 16 & 0.0 & all-linear & 32 & 1e-3 & 20 & AdamCPR & True & 1e-2 \\
    \end{tabular}
    }
\end{table}

\begin{table}[h]
    \caption{Random Found Configurations}
    \label{tab:Random-Optimizer-configs}
    \centering

    \begin{minipage}{\textwidth}
    \centering
    \textit{With batch size  = batch size 32 and gradient accumulation step [2, 4, 8].}
    \end{minipage}
    \vspace{0.5em}

    \scalebox{0.625}{
    \begin{tabular}{l|l|l|l|l|l|l|l|l|l|l|l|l|l}
    \toprule
    Dataset &  batch  & decay  & fidelity & kappa & lora & lora & lora & lora & lr & warmup & optimizer & return & weight \\
            & size & factor & & init param & alpha & dropout & layer & rank & & steps \% & & assistant mask & decay \\
    \midrule
    2407.15723v1 & 256 & 0.10 & 1 & 4.0 & 16 & 0.1 & o & 16 & 1e-5 & 10 & AdamCPR & True & 1e-4 \\
    2407.15720v1 & 64 & 0.01 & 1 & NaN & 16 & 0.0 & o & 16 & 1e-3.5 & 40 & AdamW & False & 1e-3 \\
    2407.15719v1 & 64 & 1.00 & 1 & NaN & 32 & 0.0 & o & 64 & 1e-3 & 20 & AdamW & True & 1e-1.5 \\
    2407.15708v1 & 256 & 0.01 & 1 & 4.0 & 16 & 0.0 & qkv & 32 & 1e-5 & 40 & AdamCPR & True & 1e-0.5 \\
    2407.15656v1 & 128 & 0.01 & 3 & 2.0 & 32 & 0.1 & o & 64 & 1e-3 & 10 & AdamCPR & False & 1e-2 \\
    2407.15617v1 & 64 & 0.01 & 1 & NaN & 32 & 0.1 & all-linear & 32 & 1e-3 & 10 & AdamW & True & 1e-1.5 \\
    2407.15600v1 & 64 & 0.10 & 1 & 4.0 & 32 & 0.0 & qkv & 16 & 1e-3.5 & 50 & AdamCPR & False & 1e-4 \\
    2401.04152v2 & 64 & 0.01 & 1 & NaN & 16 & 0.1 & o & 64 & 1e-3 & 10 & AdamW & True & 1e-2 \\
    \end{tabular}
    }
\end{table}
\FloatBarrier

\newpage
\section*{NeurIPS Paper Checklist}

\begin{enumerate}

\item {\bf Claims}
    \item[] Question: Do the main claims made in the abstract and introduction accurately reflect the paper's contributions and scope?
    \item[] Answer: \answerYes{}
    \item[] Justification: We create a meta-dataset from our synthetic data for transfer learning and present a novel counter-intuitive approach of finding the optimal pipeline for finetuning LLMs through transfer learning.

\item {\bf Limitations}
    \item[] Question: Does the paper discuss the limitations of the work performed by the authors?
    \item[] Answer: \answerYes{}
    \item[] Justification: Limitations can be found in appendix \ref{limits}.

\item {\bf Theory Assumptions and Proofs}
    \item[] Question: For each theoretical result, does the paper provide the full set of assumptions and a complete (and correct) proof?
    \item[] Answer: \answerNA{}
    \item[] Justification: We have empirical findings rather than theoretical results.

    \item {\bf Experimental Result Reproducibility}
    \item[] Question: Does the paper fully disclose all the information needed to reproduce the main experimental results of the paper to the extent that it affects the main claims and/or conclusions of the paper (regardless of whether the code and data are provided or not)?
    \item[] Answer: \answerYes{}
    \item[] Justification: Despite the fact that we do not publish any code, we have described our method (\ref{sec_method}) in detail. All necessary hyperparameters (\ref{ConfigurationSpace}), meta-features, tools, prompts (\ref{generation_promps}, \ref{model_evaluation}) and paper names (\ref{syn_datasets_meta}) as well as models are mentioned. 

\item {\bf Open access to data and code}
    \item[] Question: Does the paper provide open access to the data and code, with sufficient instructions to faithfully reproduce the main experimental results, as described in supplemental material?
    \item[] Answer: \answerNo{}
    \item[] Justification: We describe our method in section \ref{sec_method}. The Quick-Tune-Tool~\citep{rapant2024quick} is publicly available, to run similar experiments. The code is not executable with one click as it requires the setup of the server for the evaluation, as well as additional steps such as the setup of SSH keys and connections.

\item {\bf Experimental Setting/Details}
    \item[] Question: Does the paper specify all the training and test details (e.g., data splits, hyperparameters, how they were chosen, type of optimizer, etc.) necessary to understand the results?
    \item[] Answer: \answerYes{}
    \item[] Justification: See section \ref{sec_method} (A) and (B), Appendix \ref{ConfigurationSpace} and \ref{generation_promps} for information about the dataset and hyperparameters details.

\item {\bf Experiment Statistical Significance}
    \item[] Question: Does the paper report error bars suitably and correctly defined or other appropriate information about the statistical significance of the experiments?
    \item[] Answer: 
    \item[] Justification: In section \ref{result_sec} we state that we use standard error of the mean over eight datasets in both of our main plots \ref{abb:wall_clock} and \ref{abb:Comparison}, calculated with Seaborn \citep{Waskom2021}.

\item {\bf Experiments Compute Resources}
    \item[] Question: For each experiment, does the paper provide sufficient information on the computer resources (type of compute workers, memory, time of execution) needed to reproduce the experiments?
    \item[] Answer: \answerYes{}
    \item[] Justification: See appendix \ref{compute}.
    
\item {\bf Code Of Ethics}
    \item[] Question: Does the research conducted in the paper conform, in every respect, with the NeurIPS Code of Ethics \url{https://neurips.cc/public/EthicsGuidelines}?
    \item[] Answer: 
    \item[] Justification: Code of Ethics was observed to the best of our knowledge and belief.

\item {\bf Broader Impacts}
    \item[] Question: Does the paper discuss both potential positive societal impacts and negative societal impacts of the work performed?
    \item[] Answer: \answerNA{}
    \item[] Justification: Even if we change Quick-Tune so that we don't do Bayesian optimization, we haven't designed a new tool. We show that better performance can be achieved in the language domain. For potential positive as well as negative 
social influences, further experiments are needed. 
    
\item {\bf Safeguards}
    \item[] Question: Does the paper describe safeguards that have been put in place for responsible release of data or models that have a high risk for misuse (e.g., pretrained language models, image generators, or scraped datasets)?
    \item[] Answer: answerNA{}
    \item[] Justification: The paper proposes no risk, as the finetuned models are trained on scientific papers and will not be published either way.

\item {\bf Licenses for existing assets}
    \item[] Question: Are the creators or original owners of assets (e.g., code, data, models), used in the paper, properly credited and are the license and terms of use explicitly mentioned and properly respected?
    \item[] Answer: \answerYes{}
    \item[] Justification: The only non-own code that was used is Quick-Tune-Tools and DEHB, which is cited. The models (llama3.1 and phi3) are cited as well. Contents of arXiv e-prints are free to use for research purposes (\href{https://info.arxiv.org/help/api/tou.html}{Terms of Use for arXiv APIs}).

\item {\bf New Assets}
    \item[] Question: Are new assets introduced in the paper well documented and is the documentation provided alongside the assets?
    \item[] Answer: \answerNA{}
    \item[] Justification: The paper does not release new assets.

\item {\bf Crowdsourcing and Research with Human Subjects}
    \item[] Question: For crowdsourcing experiments and research with human subjects, does the paper include the full text of instructions given to participants and screenshots, if applicable, as well as details about compensation (if any)? 
    \item[] Answer: \answerNA{}.
    \item[] Justification: This paper does not involve crowdsourcing nor research with human subjects.

\item {\bf Institutional Review Board (IRB) Approvals or Equivalent for Research with Human Subjects}
    \item[] Question: Does the paper describe potential risks incurred by study participants, whether such risks were disclosed to the subjects, and whether Institutional Review Board (IRB) approvals (or an equivalent approval/review based on the requirements of your country or institution) were obtained?
    \item[] Answer: \answerNA{}.
    \item[] Justification: This paper does not involve crowdsourcing nor research with human subjects.

\end{enumerate}

\end{document}